\documentclass[letterpaper, 10pt, conference]{ieeeconf}  

\usepackage[most]{tcolorbox}

\IEEEoverridecommandlockouts                              

\usepackage{algorithm}

\usepackage{graphicx}
\usepackage{amsmath}
\usepackage{amssymb}
\usepackage{booktabs}
\usepackage{multirow}
\usepackage{url}
\usepackage{xcolor}
\usepackage{color}
\usepackage{pifont}
\usepackage{algpseudocode}
\usepackage{tcolorbox}

\usepackage{hyperref}
\hypersetup{pdfstartview=FitH,
            colorlinks=true,
            linkcolor=red,
            anchorcolor=blue,
            citecolor=black
            }

\newcommand{\tabincell}[2]{\begin{tabular}{@{}#1@{}}#2\end{tabular}}

\title{\LARGE \bf
Gaining the Sparse Rewards by Exploring Lottery Tickets in \\Spiking Neural Networks
}

\author{Hao Cheng$^{1*}$, Jiahang Cao$^{1*}$, Erjia Xiao$^{1}$,  Mengshu Sun$^{2}$, Renjing Xu$^{1\dagger}$
\thanks{$^*$Equal contribution; $^{\dagger}$Corresponding author.}
\thanks{$^{1}$Hao Cheng$^*$, Jiahang Cao$^*$ and Erjia Xiao are with 
        MICS Thrust, The Hong Kong University of Science and Technology (Guangzhou), 
        \newline Email: {\tt\small hcheng046@connect.hkust-gz.edu.cn,
        \newline jcao248@connect.hkust-gz.edu.cn,
        \newline exiao469@connect.hkust-gz.edu.cn
        }}%
\thanks{$^{2}$ Mengshu Sun is with Beijing University of Technology, 
         \newline Email: { \tt\small sunms@bjut.edu.cn}
        }%
\thanks{$^{1}$Renjing Xu$^{\dagger}$ is with  MICS Thrust, the Hong Kong University of Science and Technology (Guangzhou), 
        \newline Email: {\tt\small renjingxu@hkust-gz.edu.cn}
        }%
\\
}

\begin{document}

\maketitle
\thispagestyle{empty}
\pagestyle{empty}




\begin{abstract}

Deploying energy-efficient deep learning algorithms on computational-limited devices, such as robots, is still a pressing issue for real-world applications. Spiking Neural Networks (SNNs), a novel brain-inspired algorithm, offer a promising solution due to their low-latency and low-energy properties over traditional Artificial Neural Networks (ANNs). Despite their advantages, the dense structure of deep SNNs can still result in extra energy consumption. The Lottery Ticket Hypothesis (LTH) posits that within dense neural networks, there exist winning Lottery Tickets (LTs), namely sub-networks, that can be obtained without compromising performance. Inspired by this, this paper delves into the spiking-based LTs (SLTs), examining their unique properties and potential for extreme efficiency. Then, two significant sparse \textbf{\textit{Rewards}} are gained through comprehensive explorations and meticulous experiments on SLTs across various dense structures. Moreover, a sparse algorithm tailored for spiking transformer structure, which incorporates convolution operations into the Patch Embedding Projection (ConvPEP) module, has been proposed to achieve  Multi-level Sparsity (MultiSp). MultiSp refers to (1) Patch number sparsity; (2) ConvPEP weights sparsity and binarization; and (3) ConvPEP activation layer binarization. Extensive experiments demonstrate that our method achieves extreme sparsity with only a slight performance decrease, paving the way for deploying energy-efficient neural networks in robotics and beyond.

\end{abstract}    
\section{Introduction}
\label{sec:intro}

The current research in Artificial Intelligence (AI)~\cite{yang2020resolution, yang2021condensenet, zheng2023dynamic, yang2023adaptive} has entered a new stage, where the computational costs of algorithms are becoming more diverse. How to efficiently deploy corresponding algorithms in various resource-constrained scenarios, such as mobile edge devices and robotics, has become an urgent problem to solve. 
To tackle this issue, our approach analyses the problem from a dual-level perspective: (1) Redesigning existing models at the \textit{neuron level} to obtain an efficient new network structure; (2) Developing numerous sparse algorithms to reduce the parameter size of original dense models at the \textit{structure level}.

About redesigning new neurons, SNN is acclaimed as the third generation of neural networks and has increasingly gained great interest from researchers in recent years due to its distinctive properties: high biological plausibility, temporal information processing capability, and low power consumption.
Distinct from ANNs that process data in a continuous manner, SNNs operate on binary time-series data, utilizing low-power Accumulation (AC) operations over the more energy-intensive Multiply-Accumulate (MAC) operations found in ANNs. This fundamental difference not only promises significant energy savings but also aligns closely with biological neural processing. 
Additionally, SNNs follow their biological counterparts and inherit complex temporal dynamics from them, endowing SNNs with powerful abilities to extract image features in a variety of tasks, including recognition~\cite{zhou2022spikformer,deng2022temporal}, tracking~\cite{zhang2022spiking}, and images generation~\cite{cao2024spiking}. Hence, this paper leverages SNNs to address the energy-efficiency problem fundamentally.

\begin{figure*}[t]
	\setlength{\tabcolsep}{1.0pt}
	\centering
	\begin{tabular}{c}
		
		\includegraphics[width=0.8\textwidth]{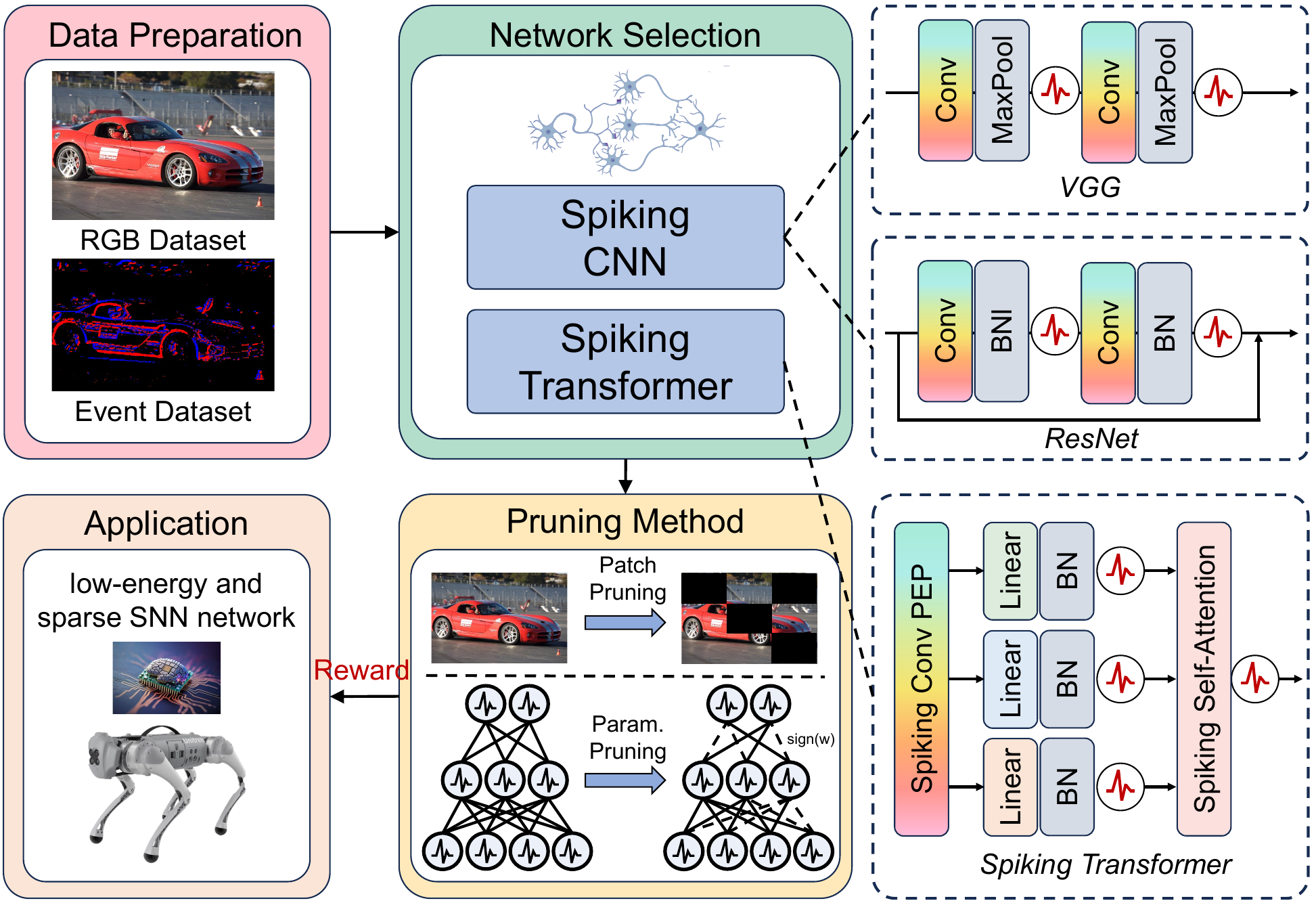} \\
		
	\end{tabular}
	\caption{The pipeline of our Spiking Lottery Tickets (SLT). The procedure begins with data preparation, utilizing either RGB or event datasets. It then progresses to the selection of network architectures, with options including CNN-based and transformer-based spiking models, where the rainbow-colored module can be sparsified by our SLT approach. The core of the process involves applying the patch pruning and parameter pruning methods, which yields rewards and returns a refined SNN network that is both energy-efficient and sparsely connected, making it ideal for implementation in resource-limited devices, e.g., robots.}
	\label{fig:main-fig}
	\vspace{-0.35cm}
\end{figure*}

However, utilizing dense SNNs could still lead to extra energy consumption. For exploring more sparse models at the structure level, recent studies have focused on applying pruning techniques in SNNs, exploring methods such as model pruning~\cite{neftci2016stochastic,rathi2018stdp,chen2021pruning, liu2023sparsespikformer}, model quantization~\cite{xu2020generative, zhou2018adaptive, fang2021deep}, and knowledge distillation~\cite{polino2018model, chen2022state}. 
However, these methods face their own set of limitations. Many are constrained to simpler models~\cite{neftci2016stochastic,rathi2018stdp}, or they lead to significant performance degradation~\cite{chen2022state, fang2021deep}, especially in more complex spiking model architectures. This gap underscores the need for more effective pruning strategies that can maintain or enhance performance while accommodating the intricate dynamics of advanced SNNs. 

To further mitigate these issues, the Lottery Ticket Hypothesis (LTH) offers a promising solution. LTH suggests that within a randomly initialized dense neural network, there exist efficient sub-networks, which can achieve the comparable accuracy of the full network within the same or fewer iterations. 
Building upon this concept, the Multi-Prize Lottery Tickets (MPLTs~\cite{diffenderfer2021multiprize}) hypothesis further refines this approach by focusing on efficient connection selection without the necessity of weight training, enhancing both weight sparseness and binarization for improved performance.


In this paper, we delve into the Lottery Tickets (LTs) in the SNN scenarios and provide comprehensive guidance for efficient sparse spiking-based methods. Although some existing work indicates the presence of LTs in SNNs~\cite{kim2022exploring, yao2023probabilistic}, they still need weight training which costs additional resources. More importantly, they have not conducted a detailed analysis of the outcomes from Spiking Lottery Tickets (SLTs) and have not further explored the properties of SLTs under Multi-Prize SLTs (MPSLTs) and Multi-level Sparse (MultiSp) conditions.
Based on this, this paper proposes an SLTs exploring algorithm for both standard CNN-based structures and transformer-based structures.
The whole pipeline of our SLT method is illustrated in Figure~\ref{fig:main-fig}. By comparing our SLTs with the original LTs, we gain the following two rewards:

\begin{tcolorbox}[colback=gray!10, arc=1mm]
\textit{\textbf{Reward 1:} 
Under the CNN-based models, 
the inner SLTs achieve higher sparsity and fewer performance losses compared to LTs.
\\
\textbf{Reward 2:} Under the transformer-based models, 
SLTs incur less accuracy loss compared to LTs counterparts at the same level of MultiSp.
}
\end{tcolorbox}


Our contributions can be summarized as follows:
\begin{itemize}
        \item We obtain two \textbf{\textit{Rewards}} that are applicable to SLTs for both CNN-based and transformer-based structures and provide a comprehensive analysis of the extreme sparsity outcomes.
         \item  We propose a Multi-level Sparsity Exploring Algorithm for spiking-based transformers. This algorithm could effectively achieve multi-level sparsity results.
        \item  We conduct extensive experiments on both RGB datasets and event datasets. Results demonstrate that our SLTs outperform the standard LTs by up to 4.58\% while achieving extreme energy savings ($>$80.0\%). 
\end{itemize}

\section{Related Works and Background}

\noindent\textbf{Spiking Neuron Networks.}
Spiking neural network is a bio-inspired algorithm that simulates the real process of signaling that occurs in brains. Compared to the artificial neural network (ANN), it transmits sparse spikes instead of continuous representations, which brings advantages such as low energy consumption and robustness. 
In this paper, we adopt the widely used Leaky Integrate-and-Fire (LIF) model~\cite{roy2019towards}, which is suitable to characterize the dynamic process of spike generation and can be defined as:
\begin{align}
    & V[n] = \beta V[n-1] + \gamma I[n] \label{eq:dis_lif1},\\
    & S[n] = \Theta (V[n] - \vartheta_{\textrm{th}})\label{eq:dis_lif2},
\end{align}
where $n$ is the time step and $\beta$ is the leaky factor that controls the information reserved from the previous time step; $V[n]$ is the membrane potential; $S[n]$ denotes the output spike which equals 1 when there is a spike and 0 otherwise; $\Theta(x)$ is the Heaviside function. When the membrane potential exceeds the threshold $\vartheta_{\textrm{th}}$, the neuron will trigger a spike and resets its membrane potential to $V_{\textrm{reset}}<\vartheta_{\mathrm{th}}$. The LIF neuron achieves a balance between computing cost and biological plausibility.

\noindent\textbf{Pruning methods in Spiking Neural Networks.}
To further improve the energy efficiency of SNN, a number of works on SNN pruning have been proposed and well-validated on neuromorphic hardware. Shi~\cite{shi2019soft} propose a pruning scheme that exploits the output spike firing of the SNN to reduce the number of weight updates during network training. Guo~\cite{guo2020unsupervised} dynamically removes
non-critical weights in training by using the adaptive online pruning algorithm. Apart from seeking the help of pruning, Rathi~\cite{rathi2018stdp} and Takuya~\cite{takuya2021training} pursue sparse SNN by using Knowledge distillation and quantization. 
Several works try to combine LTH with SNNs: Kim et al.~\cite{kim2022exploring} first investigate how to scale up pruning techniques towards deep SNNs and reveal that winning tickets consistently exist in deep SNNs across various datasets and architectures. They also propose a kind of Early-Time ticket that could alleviate the heavy search cost. Yao et al.~\cite{yao2023probabilistic} contribute a novel approach by introducing a probabilistic modeling method for SNNs. This method allows for the theoretical prediction of the probability of identical behavior between two SNNs, accounting for the complex spatio-temporal dynamics inherent to SNNs. 
\section{Method}

\noindent\textbf{Method Overview.}
The core focus of this paper lies in exploring the intrinsic properties of SLTs across different structures, including CNN-based models and transformer-based models, while comparing their performance with the original LTs across RGB and event-based datasets. For CNN-based models, we adopt the multi-prize lottery tickets hypothesis for obtaining optimal sub-networks and examine the binarized condition in detail. For transformer-based models, we investigate how extreme sparsity affects performance by integrating the parameter-level and patch-level sparsity methods and subsequently provide a thorough discussion about the balances between different sparsities. After obtaining the resulting sparse and energy-efficient networks, we could obtain two distinct rewards (mentioned in Section~\ref{sec:intro}) that benefit the implementation in resource-limited applications. The overview of our method is depicted in Figure~\ref{fig:main-fig}.

\begin{table*}[t]
    \centering
    {
    \caption{Evaluating the performance of dense ANN/SNN, original LTs/SLTs and MPLTs/MPSLTs for VGG-9/ ResNet-19 under CIFAR10, CIFAR100, DVS128Gesture and CIFAR10-DVS. }
    \label{tab:cnn}
    \small 
    \setlength\tabcolsep{8pt}
    \begin{tabular}{cccccc}
    \toprule[1.2pt]
     \multicolumn{1}{c}{\bf Architecture} & \multicolumn{1}{c}{\bf Method}   & \multicolumn{1}{c}{\bf \tabincell{c}{CIFAR10\\ Acc(\%)}} &  \multicolumn{1}{c}{\bf \tabincell{c}{CIFAR100\\ Acc(\%)}}        &  \multicolumn{1}{c}{\bf \tabincell{c}{DVS128Gesture\\ Acc(\%)}} & \multicolumn{1}{c}{\bf \tabincell{c}{CIFAR10-DVS\\ Acc(\%)}} \\
    \toprule[1.0pt] 
    \multirow{6}{1in}{\centering\emph{VGG-9} / \emph{ResNet-19}\\(2.26M) / (12.63M)} 
    & ANN  &88.10/92.05  & 68.64/76.74 & 92.92/94.83  &  78.65/80.14\\
    & SNN  & 87.71/91.78 & 67.94/75.66 &  92.78/93.27  & 78.41/80.06\\
    &LTs  &87.73/91.87  & 67.56/75.62 & 92.43/93.83  &  77.92/80.02\\
    &SLTs  & 87.67/91.66 &  65.81/74.76  & 91.73/91.77 &  77.82/79.41\\
    &MPLTs  & 87.52/90.26 & 66.74/72.01 &  87.12/89.98  & 77.01/78.66\\
    &\textbf{MPSLTs}  & \textbf{87.76/91.82} & \textbf{67.22/72.43} &   \textbf{91.70/93.22}  & \textbf{77.94/79.71} \\ 
    \bottomrule[1.2pt]
    
    \end{tabular}
    }\vspace{0.2em}
    
\end{table*}

\begin{figure*}[t]
	\setlength{\tabcolsep}{1.0pt}
	\centering
	\begin{tabular}{c}
        \includegraphics[width=1\textwidth]{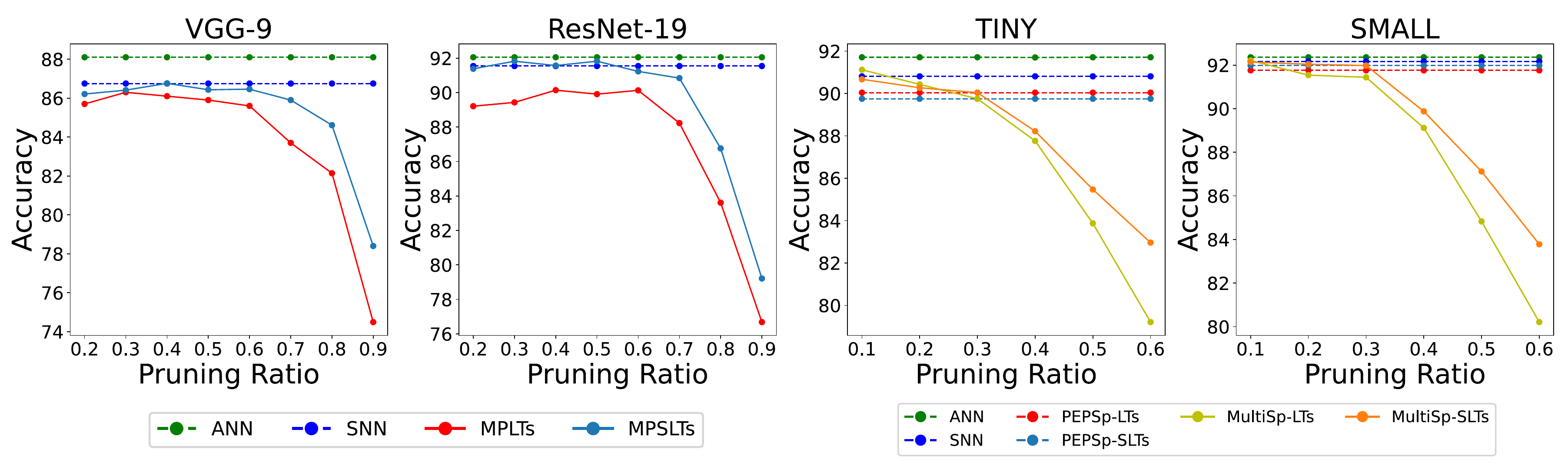}
	\end{tabular}
    \vspace{-0.35cm}
	\caption{ 
        The performance changes under different LTs and SLTs due to varying parameter-level and patch-level pruning ratios on CIFAR10.
        }
	\label{fig:c10}
	\vspace{-0.35cm}
\end{figure*}

\begin{figure*}[t]
	\setlength{\tabcolsep}{1.0pt}
	\centering
	\begin{tabular}{c}
        \includegraphics[width=1\textwidth]{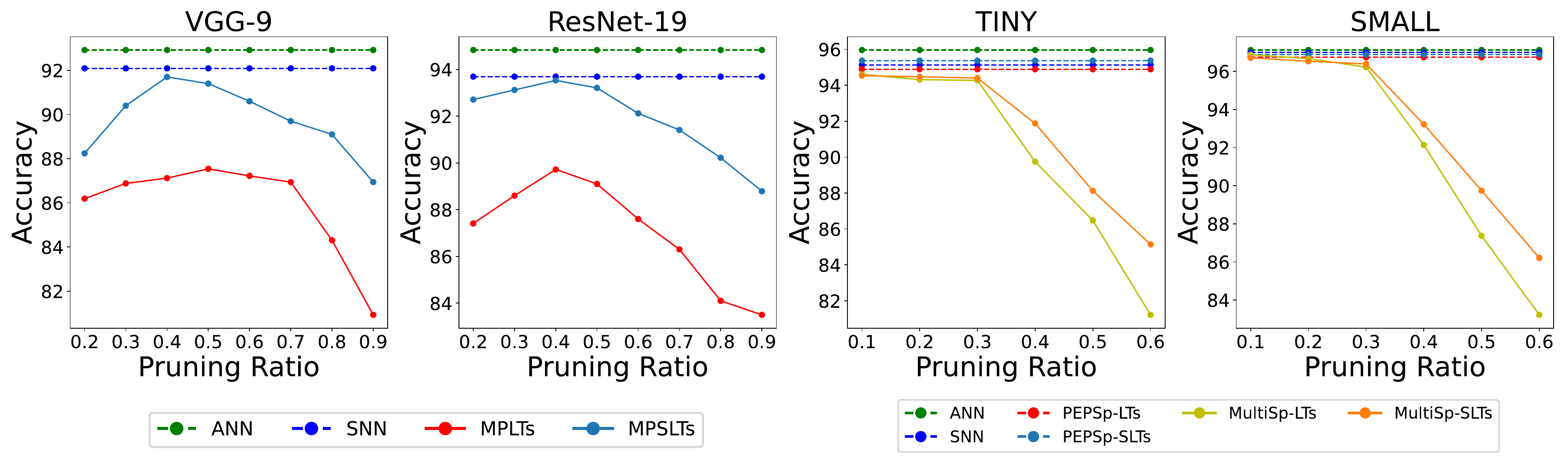}
	\end{tabular}
 \vspace{-0.35cm}
	\caption{ The performance changes under different LTs and SLTs due to varying parameter-level and patch-level pruning ratios on DVS128Gesture. 
        }
	\label{fig:dvs128}
	\vspace{-0.35cm}
\end{figure*}

\noindent\textbf{Parameters Sparse for CNN-based structures.}
Our investigation begins with the conduction of standard Lottery Tickets (LTs~\cite{ramanujan2020s}), which uncover efficient sub-networks within randomly initialized neural networks without the need for conventional weight training. Building upon this foundation, we extend our exploration to the domain of Multi-prize Lottery Tickets (MLTs~\cite{diffenderfer2021multiprize}), which introduce the added benefit of binary weight and activation representations, further enhancing the network's energy efficiency and operational effectiveness. 
As the LT process unfolds, the network's connections become increasingly sparse, driven by an evolving threshold that refines weight selection. Sub-networks emerge by retaining only those weights that surpass this threshold. Employing the strategies of LTs and MLTs allows our spiking-based CNN models to progressively achieve the sparse rewards of energy conservation and minimal performance degradation through the strategic pruning and binarization of weights.

\renewcommand{\algorithmicrequire}{\textbf{Input:}}
\renewcommand{\algorithmicensure}{\textbf{Output:}}
\begin{algorithm}[h!]
\caption{Multi-level Sparsity Exploration of Spiking Transformers}
\begin{algorithmic}[1]
\State \textbf{Input}: Spiking transformer $F(\cdot)$; ConvPEP $F_{c}(\cdot)$; Weights of ConvPEP module $W_c$; Pruning scores of ConvPEP module $S_c$; Loss function $\mathcal{L}$; Training Dataset $(P, label)$; Parameter pruning rate and epoch $\{P_a, N_a\}$; Patch pruning rate and epoch $\{P_b, N_b\}$; Patch number $n_p$; Patch embedding and Position embedding $PatE, PosE$.
\State \textbf{Output}: Return optimal binarized parameter-sparsed and patch-sparsed subnetwork $G_s(\cdot)$.
\State \textit{Randomly Initialize the weights of ConvPEP module $W_c$ and its pruning scores $S_c$.}
\State \textit{Initialize the pruning masks of ConvPEP module by $\forall m_c \in M_c, m_c = 1$, its binary subnet weights $B_c \gets \text{sign}(W_c)$ and gain term $\alpha \gets  ||M_c \odot W_c||_1 / ||M_c||_1 $.}
\For{$k=1$ to $N_{a}+N_b$}
\If{$k\leq N_a$} \hfill $\triangleright$ \small\textbf{\textit{Stage 1: param. pruning}}
    \State $S_c \gets S_c -  \eta \nabla_{S_c} \mathcal{L}(\alpha M_c \odot W_{c})$
    \State Generating the sorting indices $r_c$ and updating the pruning mask $M_c$ where its score exceeds $P_a$.
    \State Update the gain term $\alpha \gets  ||M_c \odot W_c||_1 / ||M_c||_1 $.
\State \textbf{Return} binarized sparse ConvPEP module $F_s(\cdot)$.
\ElsIf{$k> N_a$} \hfill $\triangleright$ \small\textbf{\textit{Stage 2: patch pruning}}
    \State \textit{Initialize Patch-level LTs index $id_p$}.
    \State $ PatE \gets F_{s} (P)$
    \State $id_p \gets $ Sorting  $PatE$ based on $P_b n_{p} \odot PosE$
\EndIf
\EndFor
\State  $P_{plt} \gets$ Pick the Patch-level LTs according to the $id_p\odot P$
\State \textbf{Return} $G_s(\cdot):= F(P_{plt}; M_c) \gets F(F_s(P_{plt}; \alpha(m_c \odot w_{c})))$
\end{algorithmic}
\label{Al:comb}
\end{algorithm}

\noindent\textbf{Multi-level Sparsity for transformer-based structures.}
Transformer-based models have more complex structures than CNN's, therefore, here we emphasize the sparsity of spiking transformer from two perspectives: \textit{parameter-level} and \textit{patch-level}. Regarding the sparsification of the parameters, we investigate the CNN-based patch embedding projection (ConvPEP) module where the existence of its redundancy has been proved~\cite{shendata}. MLTs methods are also utilized to achieve sparse and binarized PEP. Moreover, in the SNN-based ViT, patch-level redundancy exists, which motivates us to apply patch-pruning techniques for getting sparse rewards. 
We consider the multi-level sparsity in the spiking transformer as MultiSp, which includes 1) ConvPEP weights sparsity and binarization; 2) ConvPEP activation binarization; and 3) ConvPEP input patch number sparsity.


The Algorithm~\ref{Al:comb} presents further details of discovering spiking multi-level sparse lottery tickets. The algorithm consists of two stages,  (1) \textit{Stage 1: parameter-level pruning} and (2) \textit{Stage 2: patch-level pruning}. In the first stage, we focus on refining the weight connections and binarizing both weights and activation output. The gain term $\alpha$ needs to be continuously updated until the correct winning tickets are found. $W_c$ is the ConvPEP weights and $M_c$ is the mask that is repeatedly updating until obtaining the final Mask $M_c$ and its corresponding binarized sparse ConvPEP module $F_s(\cdot)$. 
After finding the parameter-level tickets, the updated mask $M_c$ and weights $W_c$ would be kept and start the second stage for exploring patch-level sparsity.
The original patches $P$ with different index $id_p$ from Position Embedding $PosE$ are fed into the sparse ConvPEP $F_s(\cdot)$ and obtain corresponding Patch Embedding $PatE$. 
By comparing the magnitude of output $PatE$, we can identify the top-k patches $P_{plt}$ that have the most impact on the overall performance.  
$P_{plt}$ corresponds to the patch-level SLTs of our spiking-based transformer. 
Eventually. we could obtain the resulting binarized parameter-sparsed and patch-sparsed subnetwork $G_s(\cdot)$ with valuable multi-level winning tickets.


		

\section{Experiments and Analysis}


\begin{table*}[t]
    \centering
    {
    \caption{
    The performance of ViT/Spiking-ViT, PEPSp-LTs/SLTs, MultiSp-LTs/SLTs with different Patch and Param. Pruning Ratio for TINY/SMALL under CIFAR10, CIFAR100, DVS128Gesture and CIFAR10-DVS. }
    \label{tab:vit}
 \small 
    \setlength\tabcolsep{6pt}
\begin{tabular}{cccccccc}
    \toprule[1.2pt]
\textbf{Architecture}                                            & 
\textbf{Datasets}
& ViT & Spiking-ViT  & PEPSp-LTs & PEPSp-SLTs & MultiSp-LTs & MultiSp-SLTs
\\
    \toprule[1.2pt]
\multirow{6}{*}{\begin{tabular}[c]{@{}c@{}}TINY / SMALL \\ (5M)/(22M) \end{tabular}} & \textit{Patch Pruning Ratio}                                                            & 0/0                  & 0/0                  & 0/0                                                                   & 0/0                                                                & 0.3/0.3  & \bf 0.3/0.3                                                                \\
& \begin{tabular}[c]{@{}c@{}}\textit{Param Pruning Ratio}\end{tabular}           & 0/0                 & 0/0                  & 0.4/0.4                                                                     & 0.4/0.4                                                              & 0.4/0.4      & \bf 0.4/0.4                                                                 \\ \cmidrule{2-8} 
& CIFAR10 (\%)                                                                   & 91.71/92.36               & 90.81/92.17               & 90.04/91.76                                                                  & 89.75/91.98                                     &              89.87/91.44               & \bf 90.21/91.98                                                               \\
& CIFAR100 (\%)                                                                 & 74.37/75.12               & 73.84/75.43               & 74.31/75.22                                                                  & 73.68/74.86                                                 &         73.21/74.09        & \bf 73.82/74.89                                                               \\
& DVS128Gestrure  (\%)                                                          & 95.96/97.12              & 95.13/96.99               & 94.89/96.74                                                                  & 95.37/96.88                                               &          94.27/96.22         & \bf 94.40/96.39                                                               \\
& CIFAR10-DVS (\%)                                                              & 78.43/80.01               &  77.86/79.77               & 76.61/78.42                                                                 & 77.14/78.91                                        &           76.69/78.02               & \bf 77.03/78.43                    \\                                           
    \bottomrule[1.2pt]
\end{tabular}
    }\vspace{0.2em}
\end{table*}

\begin{figure*}[t]
	\setlength{\tabcolsep}{1.0pt}
	\centering
	\begin{tabular}{c}
        \includegraphics[width=1\textwidth]{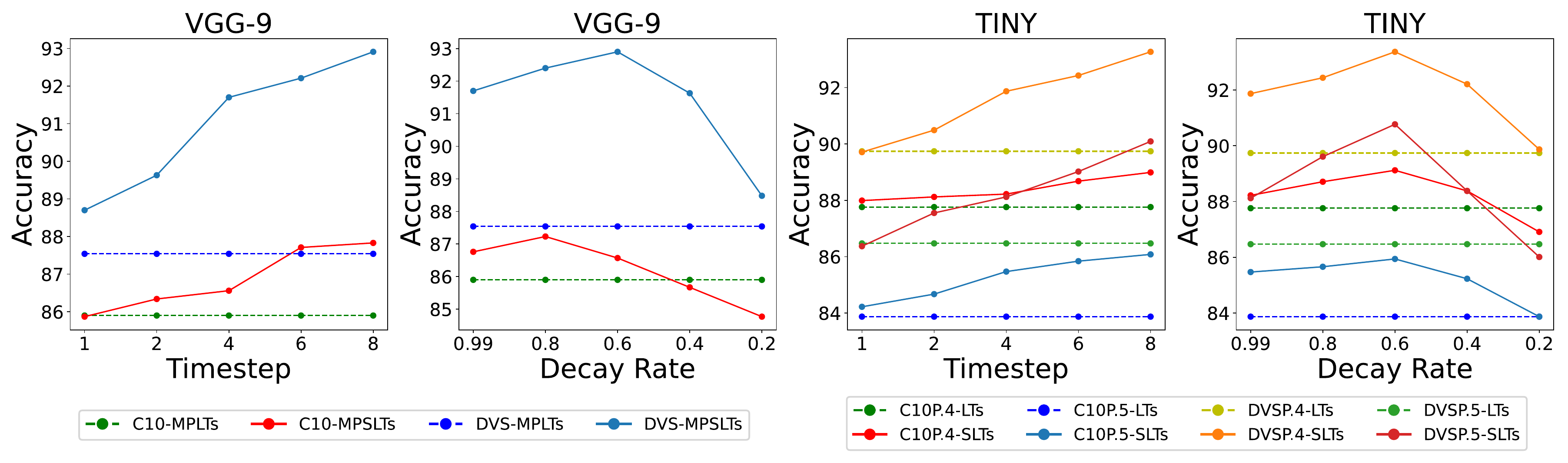}
	\end{tabular}
 \vspace{-0.35cm}
	\caption{ The performance effect under different Timestep and Decay Rate to VGG-9 and TINY in CIFAR10 (C10) and DVS128Gesture (DVS). The P.4 and P.5 indicate the parameter Pruning Ration is 0.4 and 0.5.
        }
	\label{fig:TDR}
	\vspace{-0.35cm}
\end{figure*}


\subsection{Experimental Setting}
In this paper, 
to verify our finding rewards and proposed Algorithm~\ref{Al:comb},
we use two CNN-based structures VGG-9 and ResNet-19, and two transformer-based structures TINY and SMALL referring to the model scale of DeiT-Tiny and Deit-Small~\cite{touvron2021training}. The spiking version of the above structures are from~\cite{fang2021deep,zhou2022spikformer}. We adopt two RGB-based datasets CIFAR10 and CIFAR100, two event-based datasets DVS128Gesture~\cite{amir2017low} and CIFAR10-DVS~\cite{cheng2020structure}.
When exploring parameter sparse LTs and SLTs in CNN modules, we use pruning ratio $P_a \in \{ 0.2, 0.3, 0.4, 0.5, 0.6, 0.7, 0.8\}$,
$P_b\in \{0.1, 0.2, 0.3, 0.4, 0.5, 0.6\}$ is the patch-level pruning ratio for the input patch number of ConvPEP.
For the parameters of SNN, we adopt the most commonly used LIF neuron with timestep $T=4$ and decay rate $\lambda=0.99$ in our main experiments. 
Additionally, in order to explore the specific relationship between various SNN component parameters and spiking lottery tickets, we 
further use time step $T\in \{1,2,4,6,8\}$ and decay rate $\lambda\in \{0.99, 0.8, 0.6, 0.4, 0.2\}$. We use the Adam optimizer with a base learning rate of $0.1$ to facilitate the learning process. Surrogate gradient training methods~\cite{neftci2019surrogate} for SNN are adopted. The models for conducting experiments are implemented based on Pytorch and SpikingJelly~\cite{SpikingJelly}.
Since our main focus is finding the best spiking lottery tickets while maintaining accuracy, we only use the most primitive direct training without any training tricks in all our experiments. Furthermore, the Appendix~\ref{sec:appendix} includes more detailed interpretations.

\subsection{General Performance Analysis}
\begin{figure*}[t]
	\setlength{\tabcolsep}{1.0pt}
	\centering
	\begin{tabular}{c}
        \includegraphics[width=1\textwidth]{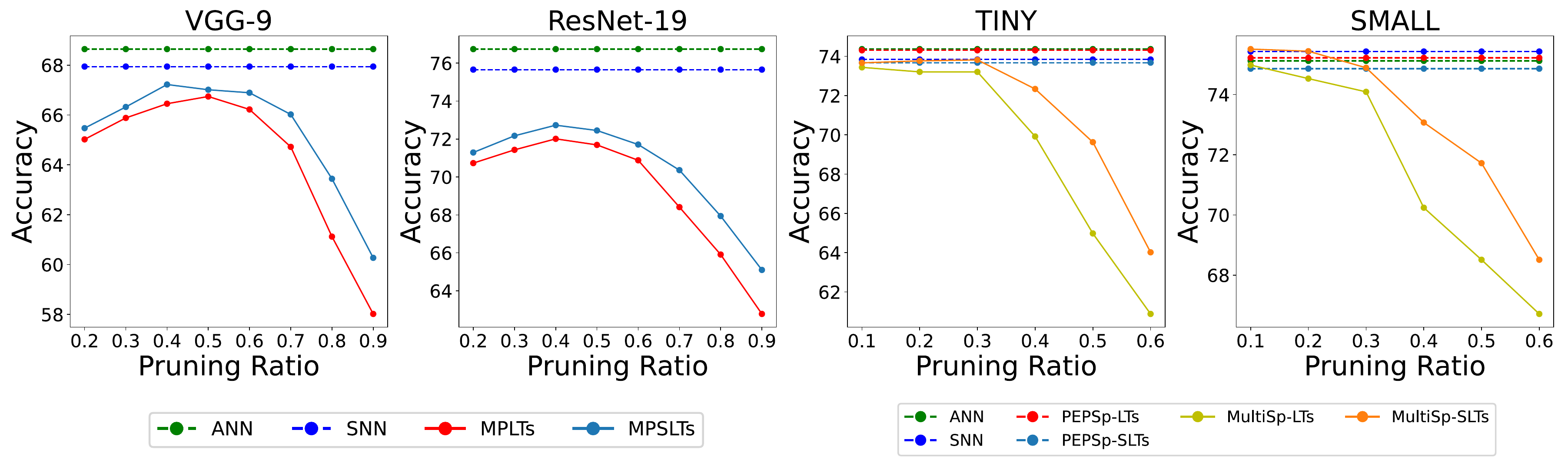}
	\end{tabular}
	\caption{ The performance changes under different LTs and SLTs due to varying parameter-level and patch-level pruning ratios on CIFAR100.
        }
	\label{fig:figdecay}
	\vspace{-0.35cm}
 \label{fig:c100}
\end{figure*}

\begin{figure*}[t]
	\setlength{\tabcolsep}{1.0pt}
	\centering
	\begin{tabular}{c}
        \includegraphics[width=1\textwidth]{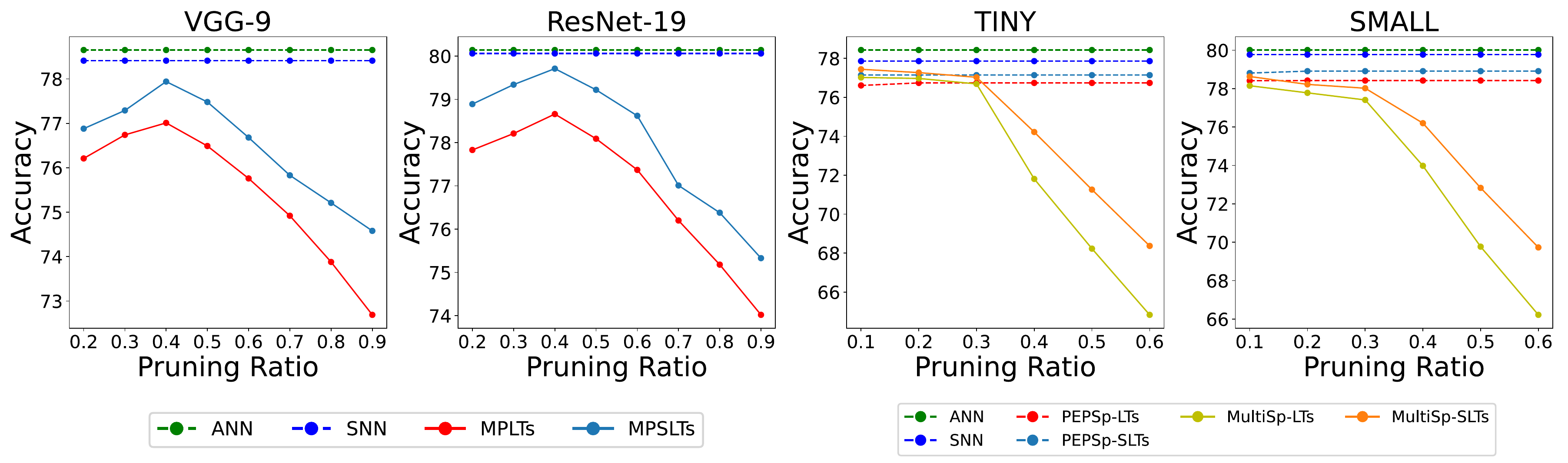}
	\end{tabular}
	\caption{ The performance changes under different LTs and SLTs due to varying parameter-level and patch-level pruning ratios on CIFAR10DVS.
        }
	\label{fig:c10dvs}
	\vspace{-0.35cm}
\end{figure*}

To comprehensively explore the inherent characteristics of spiking LTs and compare them with ANN LTs, Figure~\ref{fig:c10},~\ref{fig:dvs128},~\ref{fig:c100}, and~\ref{fig:c10dvs} respectively evaluate our used four structures on four different datasets under different pruning ratios $P_a$ and $P_b$. 

\noindent\textbf{CNN-based Sparsity.} In the study focusing on structures with convolution operations, based on the left two sub-figures of the aforementioned four figures, we first investigate the performance of Multi-prize LTs (MPLTs) and SLTs (MPSLTs) under CNN-based structures across different values of $P_a$. Additionally, we compare their performance with the corresponding original dense ANNs and SNNs.
Through the observation, we can conclude that MPSLTs outperform MPLTs across different structures and types of datasets. This observation supports the \textbf{\textit{Reward 1}} mentioned in the Section~\ref{sec:intro}.
In addition, the optimal winning tickets in both MPLTs and MPLTs occur at the pruning ratio of $P_a=0.4$. Therefore, in the subsequent exploration regarding patch-level sparsity, we will fix the $P_a=0.4$ for the ConvPEP.

\noindent\textbf{Transformer-based Sparsity.}
We attempt to explore the performance of Multi-level LTs (MultiSp-LTs) and SLTs (MultiSp-SLTs) in spiking transformers under different $P_b$. We also compare the ConvPEP sparse LTs (PEPSp-LTs) and SLTs (PEPSp-SLTs) which only perform the parameter-level sparsity in the ConvPEP modules serving as baselines. As illustrated in the right two sub-figures,  
both MultiSp-LTs and MultiSp-SLTs can maintain similar performance as PEPSp-LTs and PEPSp-SLTs at $P_b=0.3$.
However, during the descent process, MultiSp-SLTs exhibit smaller performance degradation compared to MultiSp-SLTs, indicating that MultiSp-SLTs possess better sparse robustness. This aligns with the \textbf{\textit{Reward 2}} proposed above.

\noindent\textbf{Overall Results.}
To summarize the best-performing MPLTs and MPSLTs under different CNN-based structures, 
we present Table~\ref{tab:cnn}. When $P_a=0.4$, the MPSLTs with the best performance under VGG-9/ResNet-19 on CIFAR10, CIFAR100, DVS128Gesture and CIFAR10-DVS yield accuracy improvement with 
$0.24/1.56$, $0.48/0.42$, $4.58/3.24$ and $0.93/1.05$ ($\%$) compared to MPLTs.
It is important to note that MPSLTs gain more performance increases under the event-based dataset compared to the RGB dataset.
Finally, regarding the results of standard SNN/ANN transformers, PEPSp-LTs/PEPSp-SLTs and MultiSp-LTs/MultiSp-SLTs, we select the best-performing results with the most suitable sparse parameters, namely $\{P_a; P_b\}=\{0.4,0.3\}$, and summarize them in Table~\ref{tab:vit}. Our Multip-SLTs also achieve better outcomes than the ANN's version, demonstrating the effectiveness of our spiking-based sparsity method.


\begin{figure*}[t]
\centering
	\setlength{\tabcolsep}{1.0pt}
	\centering
	\begin{tabular}{c}
        \includegraphics[width=1\textwidth]{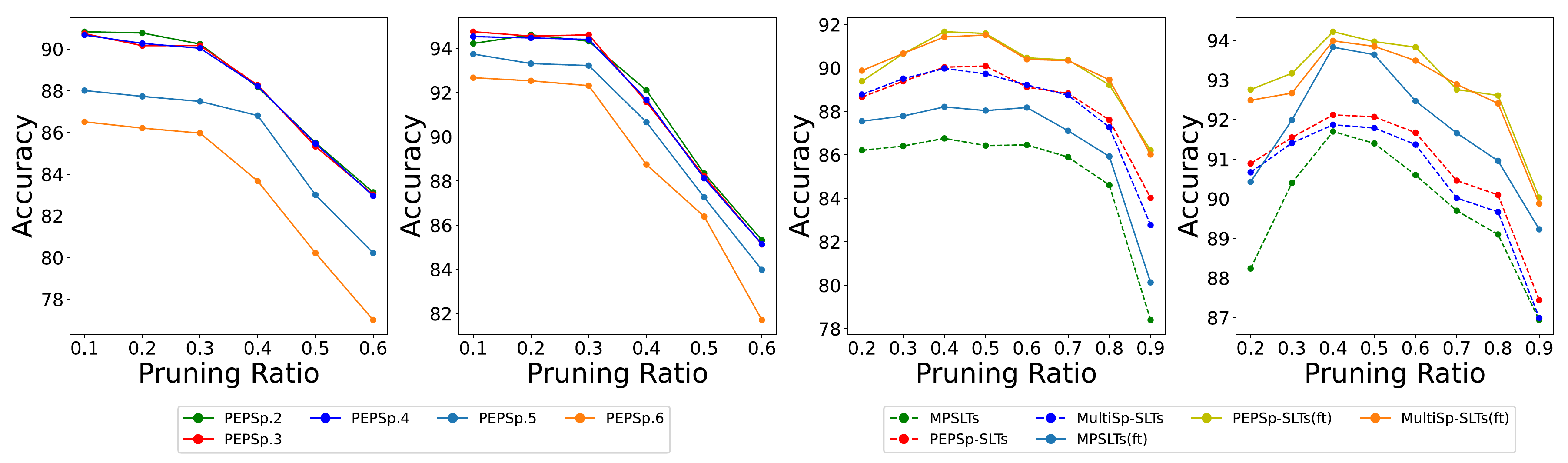}
	\end{tabular}
	\caption{ The left two figures: The impact of different param. pruning ratios ($0.2, 0.3, 0.4, 0.5, 0.6$) on overall performance of patch-lever sparsity. The right two figures: The Fing-Tuning (FT) effect to MPTSLTs, and PEPSp-SLTs and MultiSp-SLTs. 
        }
	\vspace{-0.35cm}
 \label{fig:AbStudy}
\end{figure*}
\subsection{Analysis on SNN-related Parameters}
\begin{figure*}[t]
	\setlength{\tabcolsep}{1.0pt}
	\centering
	\begin{tabular}{c}
        \includegraphics[width=1\textwidth]{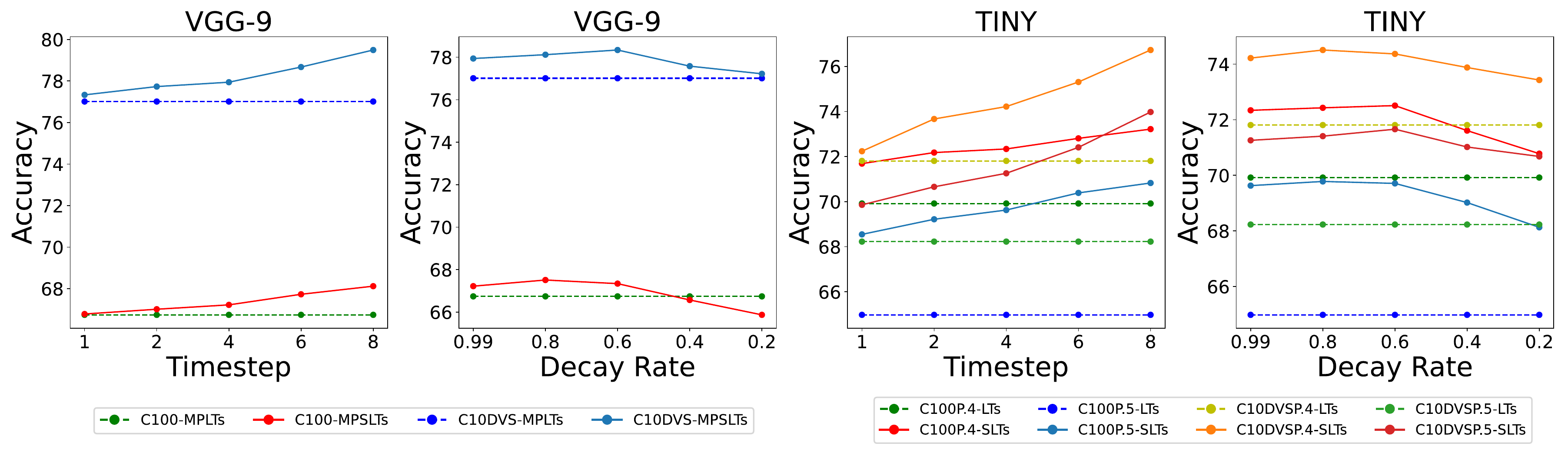}
	\end{tabular}
	\caption{ The performance effect under different Timestep and Decay Rate to VGG-9 and TINY in CIFAR100 (C100) and CFIAR10-DVS (C10DVS). The P.4 and P.5 indicate the parameter Pruning Ration is 0.4 and 0.5.
        }
	\label{fig:TDR_L}
	\vspace{-0.35cm}
\end{figure*}

In this section, we clarify the impact of the unique timestep $T$ and decay rate $\lambda$ in SNNs on the performance under different types of SLTs
In Figure~\ref{fig:TDR} and Figure~\ref{fig:TDR_L}, we select $T\in \{1,2,4,6,8\}$ and $\lambda\in \{0.99, 0.8, 0.6, 0.4, 0.2\}$ as our parameter variables.
We conduct experiments with VGG-9 and TINY on four datasets.
Specifically for VGG-9, we demonstrate the performance variation of MPSLTs under the modification of $T$ and $\lambda$. 
Under the TINY, we select patch-level pruning ratio $P_b=0.4/0.5$ (P.4/P.5), which performance just begins to decline in terms of patch-level sparsity. We then illustrate the performance variation of MultiSp-SLTs under different $T$ and $\lambda$ compared with MultiSp-LTs (abbreviated as SLTs/LTs in Figure~\ref{fig:TDR} and Figure~\ref{fig:TDR_L}).

\noindent\textbf{Time Step.} 
The MPSLTs of VGG-9 and the MultSp-SLTs on transformer show performance improvement with the increase of timestep $T$, also maintaining the properties of the two rewards discovered above.

\noindent\textbf{Decay Rate.}
The performances of MPSLTs of VGG-9 and the MultSp-SLTs on transformer do not simply exhibit a monotonic increase or decrease with the value change of decay rate. Instead, it shows an increase followed by a decrease as $\lambda$ decreases. This implies that in our pursuit of spiking-based MultiSp-SLTs in the future, we need to first make a more detailed selection of decay rate $\lambda$ values.

\subsection{Comparisons of Power Consumption}
To better understand the effects of our spiking LTs on
energy using, we estimate the theoretical power consumption on neuromorphic chips (detailed in Appendix~\ref{subsec:theor_energy}). 
We compute the energy results of VGG-9 in CIFAR10, where the MPSLTs only cost 0.079 $mJ$ per image, exhibiting merely 11.2\% (0.079/0.708) energy consumption compared to its ANN counterpart. In addition, our MPSLT further reduces energy consumption by 15.1\% on top of the dense SNN, i.e., 0.079 $mJ$ \textit{vs.} 0.093 $mJ$. This \textbf{extreme energy-savings} further validates the possibility of deploying our algorithms on resource-constrained machines in the future.

\begin{figure}[htbp]
	\setlength{\tabcolsep}{1.0pt}
	\centering
	\begin{tabular}{c}
        \includegraphics[width=0.7\linewidth]{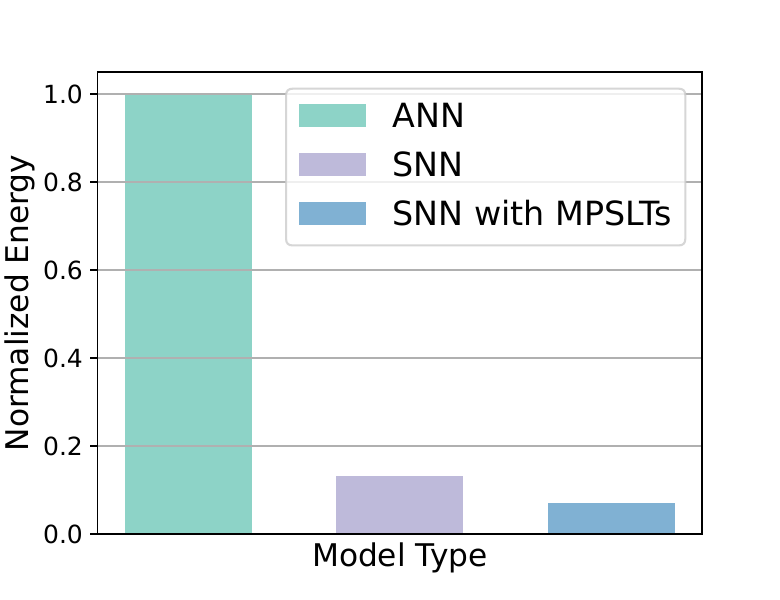}
	\end{tabular}
 	\vspace{-0.25cm}
	\caption{Comparison of energy consumption using (a) ANN, (b) SNN without any pruning, and (c) SNN with MPSLTs. Our method achieves extreme energy-saving with only 0.079 $mJ$/image.  
        }
	\label{fig:energy_bar}
 	\vspace{-0.2cm}
\end{figure}

\subsection{Ablation Study}

\noindent\textbf{Evaluation of the Balance Between $P_a$ and $P_b$.}
Since the MultiSp-SLTs in transformer-based models represent a multi-level combination of ConvPEP parameter-level sparsity (PEPSp) and patch-level sparsity, the modification in different pruning ratios $P_a$ and $P_b$ are bound to influence each other. 
In this section, we further validate the performance variation of PEPSp with additional $P_a\in \{0.2,0.3,0.5,0.6\}$ (abbreviated as Sp.x)
under different values of $P_b$.
The two left sub-figures of Figure~\ref{fig:AbStudy} illustrate that the performance trends of overall patch-level sparsity are similar when executing $P_a$ modification. However, for overall performance under different $P_b$, there exists a specific threshold between $P_a=0.4$ and $P_a=0.5$. A noticeable decline in overall performance would occur once $P_a$ exceeds this threshold.

\noindent\textbf{Effect of Fine-tuning Strategy.}
Since the LTs pruning methods do not train the weights directly, therefore, we hope to explore how fine-tuning can further affect the final performances.
We conduct fine-tuning (FT) separately for MPSLTs, PEPSp-SLTs, and MultiSp-SLTs. As shown in the right sub-figures in Figure~\ref{fig:AbStudy}, the results demonstrate that SLTs would generate performance improvements through additional fine-tuning strategy.

\section{Conclusion}


In summary, this paper tackles the pressing need for energy-efficient algorithms by exploring the extreme sparsity in low-energy spiking neural networks. In detail, we focus on considering the existence of lottery tickets in SNNs and their corresponding unique properties compared with ANN lottery tickets. Two valuable sparse rewards are investigated under the spiking CNN-based and transformer-based models. In addition, we propose a sparse algorithm tailored to the spiking transformer, which incorporates convolution operations into the Patch Embedding Projection (PEP) module, achieving multi-level sparsity. Comprehensive experiments show that our approach results in significant sparsity while minimally impacting performance, thereby facilitating the implementation of energy-efficient neural networks across robotics and additional fields.



{\small
\bibliographystyle{IEEEtran}
\bibliography{IEEEexample}
}





\section{Appendix}
\label{sec:appendix}
\subsection{Theoretical Power Consumption}
\label{subsec:theor_energy}
To calculate the theoretical energy consumption, we begin by determining the synaptic operations (SOPs). The SOPs for each block can be calculated using~\cite{zhou2022spikformer}: 
\begin{equation}
    \operatorname{SOPs}(l)=fr \times T \times \operatorname{FLOPs}(l)
\end{equation}
where $l$ denotes the block number in the model, $fr$ is the firing rate of the input spike train of the block and $T$ is the time step of the spike neuron. $\operatorname{FLOPs}(l)$ refers to floating point operations of $l$ block. 

To estimate the theoretical energy consumption of the spiking model, we assume that the operations are implemented on a $45 nm$ hardware, with energy costs of $E_{MAC} = 4.6 pJ$ and $E_{AC} = 0.9 pJ$, respectively. 
According to~\cite{yao2023attention}, the calculation for the theoretical energy consumption of SNN is given by:
\begin{align}
    E_{SNN} 
     &= E_{flop}  \times {\rm FLOP}^1_{{\rm Conv}} \nonumber \\  
     &+ E_{sop} \times \left(\sum_{n=2}^{N}{\rm SOP}^n_{{\rm Conv}} + \sum_{m=1}^{M}{\rm SOP}^m_{{\rm FC}}\right) 
     \label{eq:flop}
\end{align}
where $N$ and $M$ denote the number of spiking convolutional layers and spiking linear layers. 
We first sum up the SOPs of all $Conv$ layers (except the first layer), and $FC$ layers and multiply by $E_{flop}$. For the first convolutional layer of SNN, we calculate the energy consumption utilizing FLOPs due to the spike encoding operation performed here.

\end{document}